\documentclass[a4paper,twoside]{article}

\usepackage{epsfig}
\usepackage{subcaption}
\usepackage{calc}
\usepackage{amssymb}
\usepackage{amstext}
\usepackage{amsmath}
\usepackage{amsthm}
\usepackage{multicol}
\usepackage{pslatex}
\usepackage{apalike}
\usepackage{algorithm2e}
\usepackage[bottom]{footmisc}

\usepackage{cite}
\usepackage{amsmath,amssymb,amsfonts}
\usepackage{algpseudocode}
\usepackage{url}
\usepackage{graphicx}
\usepackage{textcomp}
\usepackage{multicol, blindtext}
\usepackage{xcolor}
\usepackage{hyperref}
\usepackage{booktabs}
\usepackage{textgreek}
\usepackage{multirow}
\usepackage[normalem]{ulem}
\hypersetup{
    colorlinks=false,
    linkcolor=blue,
    filecolor=magenta,      
    urlcolor=blue,
}
\usepackage{orcidlink}
\def\BibTeX{{\rm B\kern-.05em{\sc i\kern-.025em b}\kern-.08em
    T\kern-.1667em\lower.7ex\hbox{E}\kern-.125emX}}

\usepackage{SCITEPRESS}     

\newcommand{\PREPRINTYEAR}{2024}
\newcommand{\PUBLISHEDIN}{International Conference on Informatics in Control, Automation and Robotics (ICINCO)}

\usepackage[placement=top,vshift=-7]{background}
\SetBgScale{1.0}
\SetBgContents{\PUBLISHEDIN. PREPRINT VERSION - DO NOT DISTRIBUTE. 
}
\SetBgColor{black}
\SetBgAngle{0}
\SetBgOpacity{1.0}

\begin{document}

\thispagestyle{empty}
\onecolumn
{
  \topskip0pt
  \vspace*{\fill}
  \centering
  \LARGE{%
    \copyright{} \PREPRINTYEAR~\PUBLISHEDIN\\\vspace{1cm}
    Personal use of this material is permitted.
    Permission from \PUBLISHEDIN~must be obtained for all other uses, in any current or future media, including reprinting or republishing this material for advertising or promotional purposes, creating new collective works, for resale or redistribution to servers or lists, or reuse of any copyrighted component of this work in other works.}
    \vspace*{\fill}
}
\NoBgThispage
\twocolumn          	
\BgThispage

\title{Intuitive Human-Robot Interface: A 3-Dimensional Action Recognition and UAV Collaboration Framework}

\author{Akash Chaudhary$^{1}$\orcidlink{0000-0001-7857-7641}, Tiago Nascimento$^{1,2}$\orcidlink{0000-0002-9319-2114} and Martin Saska$^{1}$\orcidlink{0000-0001-7106-3816}
\affiliation{\sup{1}Faculty of Electrical Engineering, Czech Technical University in Prague, Technicka 2, 166 27 Prague, Czech Republic}
\affiliation{\sup{2}Universidade Federal da Paraíba, Brazil}
\email{\{chaudaka, pereiti1, martin.saska\}@fel.cvut.cz}
}


\abstract{Harnessing human movements to command an Unmanned Aerial Vehicle (UAV) holds the potential to revolutionize their deployment, rendering it more intuitive and user-centric. In this research, we introduce a novel methodology adept at classifying three-dimensional human actions, leveraging them to coordinate on-field with a UAV. Utilizing a stereo camera, we derive both RGB and depth data, subsequently extracting three-dimensional human poses from the continuous video feed. This data is then processed through our proposed k-nearest neighbour classifier, the results of which dictate the behaviour of the UAV. It also includes mechanisms ensuring the robot perpetually maintains the human within its visual purview, adeptly tracking user movements. We subjected our approach to rigorous testing involving multiple tests with real robots. The ensuing results, coupled with comprehensive analysis, underscore the efficacy and inherent advantages of our proposed methodology.}

\onecolumn \maketitle \normalsize \setcounter{footnote}{0} \vfill

\section{Introduction}
\label{sec:intro}

In the rapidly evolving field of robotics, intuitive human-robot interaction (HRI) remains a pivotal challenge. The ability for robots to accurately interpret and respond to human actions is crucial for advancing their integration into diverse applications, from industrial automation \cite{Vysocky2016} and healthcare \cite{Mohebbi2020} to agriculture \cite{Vasconez2019} and autonomous vehicles \cite{Mokhtarzadeh2018}. Traditional control interfaces, such as joysticks and remote controllers, often fail to provide the natural, seamless interaction that users require. This gap underscores the need for more intuitive and user-friendly methods to enhance human-robot collaboration, particularly in the context of Unmanned Aerial Vehicles (UAVs).

\begin{figure}[!t]
\vspace{0.3cm}
\centerline{\includegraphics[width = 0.4\textwidth]{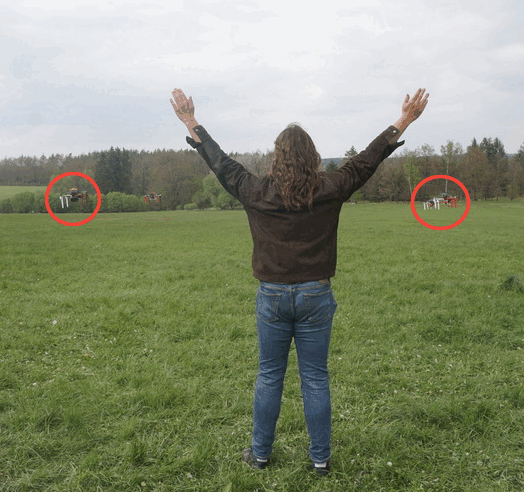}}
\caption{A Group of UAVs controlled by a human operator in an open field.}
\label{fig:introswarm}
\end{figure}

Recent advancements in action recognition and human-robot collaboration have shown significant promise in addressing these challenges. For instance, research on annotating human actions in 3D point clouds has demonstrated the importance of precise and flexible data for collaborative tasks with industrial robots, emphasizing the potential of 3D data to improve HRI systems \cite{Novel2023}. Similarly, the integration of natural language instructions and 3D gesture recognition has enhanced the intuitiveness of human-robot interaction, making it more effective for industrial applications by facilitating a more natural communication interface \cite{Natural2024}. Studies focusing on end-to-end systems for human-UAV interaction highlight the relevance of intuitive control mechanisms in field applications, showing how 3D gestures can be effectively used to control UAVs in real-time scenarios \cite{Intuitive2020}. Additionally, the use of intuitive interaction systems, such as RFHUI, highlights the significance of gesture recognition in enhancing the ease of operation and control of UAVs in 3D space \cite{RFHUI2018}.

Despite these advancements, interpreting human gestures and translating them into robotic actions remain significant hurdles. The complexity of human movements and the variability in their execution pose challenges for robotic systems, particularly those with limited computational power, such as UAVs. Our research aims to bridge this gap by proposing a novel methodology for real-time, low computationally expensive, three-dimensional action recognition and UAV collaboration. By leveraging stereo cameras to capture both RGB and depth data, we can extract and classify three-dimensional human poses from continuous video feeds. This approach enables the UAV to accurately interpret human movements and respond appropriately \ref{fig:introswarm}, thereby enhancing the intuitiveness and effectiveness of human-UAV interaction. Our main contributions are:

\begin{enumerate}
    \item A new method to estimate three-dimensional full-body pose from available 2D poses.
    \item A proposed feature vector space tailored for Human Motion Recognition.
    \item A unique, fast and lightweight human motion classifier suitable for UAVs with limited computing power.
\end{enumerate}

\section{Related Works}
\label{sec:relatedworks}

Human-robot interaction (HRI) offers a myriad of methodologies. Among these, the most intuitive is the teleoperation of a robot through a physical controller. Yamada et al.\cite{Yamada2015} employ this strategy by integrating it with virtual reality to direct robots in construction scenarios. Conversely, Sathiyanarayanan et al.\cite{Sath2015} harness a wearable armband, translating its gestures into commands for robot systems. For individuals with disabilities, voice-controlled systems present an invaluable solution. Gundogdu et al.\cite{Gind2018} pioneered such a system, facilitating the operation of prosthetic robot arms.
Our proposed action classification method adeptly amalgamates RGB video data with depth video output, enabling the classification of 3D human movements. A prior research \cite{Akash2022} explored a similar domain, but was constrained to 2D data, thereby limiting the dominion of the user over the robot and impeding optimal performance. Our tailored feature space for k-Nearest Neighbor further enhances its effectiveness even amidst intricate actions by using depth information, a custom feature space, and fast lookup times during the classification process.

In the realm of sequence classification for human motion classification, Celebi et al. \cite{Celebi2013} marked a significant advancement by introducing a weighted Dynamic Time Warping (DTW) methodology that achieved a remarkable accuracy of 96\%, a substantial improvement from the preceding state-of-the-art's 62.5\%. Following this, Rwigema et al. \cite{Rwigema2019} refined the system by integrating a differential evolution strategy to optimize DTW's weightings. This enhanced approach achieved a stellar accuracy of 99.40\%. However, its extended processing time posed challenges for real-time classification applications. In parallel, Schneider et al. \cite{Schneider2019} melded DTW with the one-nearest neighbour technique for movement classification. Their methodology closely aligns with our approach, prompting a comparative analysis between their method and ours to discern our method's efficacy. Additionally, Yoo et al. \cite{Yoo2022} demonstrated rapid processing capabilities in their classification system, although it necessitated an auxiliary IMU sensor for optimal performance. Additionally, their system is limited to palm-action classification. Our proposed approach endeavours to address the trifecta of challenges: processing speed, accuracy, and sensor dependency, offering a holistic solution in the domain of human motion recognition.

\section{Methodology}
\label{sec:method}
Our overarching ambition is to architect a method that is apt for deployment on flying robots and proficient in human detection, action classification, and the subsequent translation of these actions into robotic tasks. Given the inherent computational limitations of UAVs, the challenge lies in devising a classifier that synergizes accuracy with computational efficiency.

A salient feature of our methodology is its prowess in classifying 3D actions. This capability augments the spectrum of detectable movements, offering an enriched, intuitive user interaction. It not only mitigates potential classification errors where 2D projections might be misleading but also empowers the system to discern directional nuances from actions, granting users refined control over the trajectory of the robot.

\begin{figure*}[ht]
\vspace{0.3cm}
\centerline{\includegraphics[width = 0.8\textwidth]{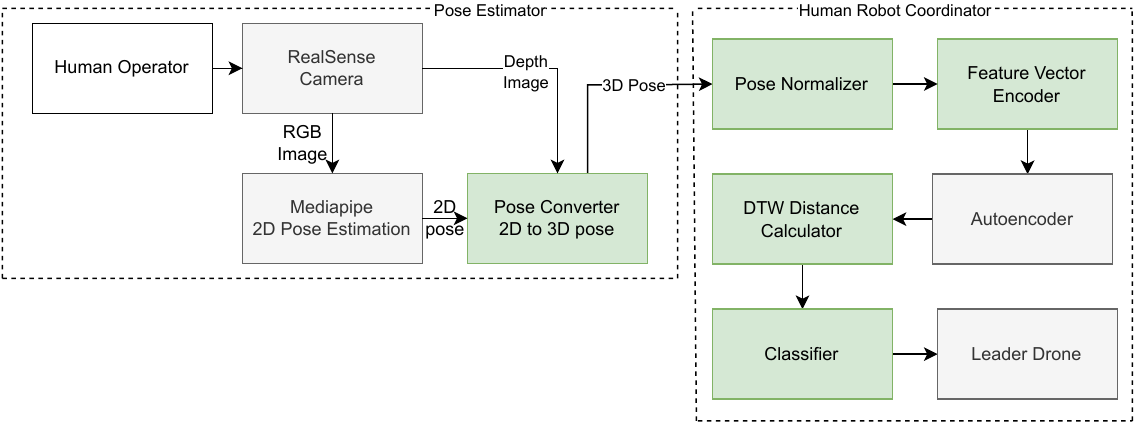}}
\caption{Flowchart depicting the action classifier and its use in UAV control, with the green blocks depicting our contribution. }
\label{fig:systemgraph}
\end{figure*}

Structured methodically, our approach is segmented into three core modules: Pose Estimation, Action Classification and UAV Control. An overview of our proposed approach can be seen in Fig. \ref{fig:systemgraph}.

\section{Pose Estimation}
\label{sec:poseest}
The recognition of static gestures can be achieved either directly from RGB images via neural networks, or by distilling skeletal data and applying subsequent functions to this data for classification. For the 3D representation, we initially harness a pre-existing model to derive 2D poses from live video and subsequently use depth images to gauge the distance of the pose joints from the camera. 

\subsection{2D Pose Estimation}
\label{sec:2dpose}
Our choice was to use MediaPipe, an implementation rooted in BlazePose \cite{Bazarevsky2020}. This Google-based pose estimator is adept at extracting 2D pose information from an RGB image in a live video stream, capturing 33 key landmarks of human anatomy. Calibration in MediaPipe is anchored on the 'Vitruvian Man', ensuring accurate scale and orientation recognition. This precision is paramount- especially in aerial robotics, where UAV movement can result in a non-horizontal camera orientation relative to the ground. 

MediaPipe operates in a distinct fixed-rate loop, processing the concurrent image. If image callbacks outpace the MediaPipe loop rate, some frames might be skipped. Despite this, our onboard system managed a commendable 30 Hz MediaPipe loop frequency, ensuring near-real-time responsiveness.

\subsection{3D Pose Estimation}
\label{sec:3dpose}
Our methodology leverages 2D poses to predict the z-coordinate for each landmark. Assuming the camera plane as the X-Y plane, each body pose keypoint corresponds to a coordinate \((x,y)\). Beyond just RGB, the Intel Realsense also provides depth images, albeit with a slight misalignment. By aligning the depth image with the RGB counterpart, we ensure a precise correspondence between them.

The z-coordinate is deduced by overlaying the 2D poses on the depth image. An area surrounding each landmark is considered, factoring in the surrounding pixels for the best z-coordinate estimate. The area of consideration inversely correlates with the distance from the camera. This necessitates an initial estimation of this distance, achieved by superimposing the coordinates of the shoulders and hips on the depth image, creating a bounding quadrilateral. Given the potential for background inclusion, we opt for the first quartile average of the depth values, excluding background measurements. 

Given the output from an RGB-D camera like Realsense, each pixel's correspondence to an actual physical area varies with the distance. At a distance of 1 meter, each pixel represents an area of \(1.5\text{mm} \times 1.5\text{mm}\), whereas at 6 meters, it is \(9\text{mm} \times 9\text{mm}\). With human proportions in consideration, we scale the number of pixels to match a \(90\text{mm} \times 90\text{mm}\) area. The subsequent z-coordinate estimation, combined with the MediaPipe output, yields a comprehensive 3D pose.

\section{Full-Body Action Classification}
\label{sec:actionclas}
Given the computational constraints on our fleet of UAVs, the emphasis is on a lightweight algorithm that retains accuracy. A balance is struck with a custom k-nearest Neighbor (kNN) classifier. Our approach to action classification is holistic, beginning with the careful design of a feature space. By leveraging the relationship between several joints in 3D space and in time, we derive a representative feature vector that encapsulates the dynamics of human posture. The core of our methodology adopts a two-stage classification approach. In the initial stage, Feature embedding translates pose landmarks into meaningful vector space. Autoencoders are then employed to reduce the dimensionality of the embeddings. Subsequently, Dynamic Time Warping (DTW) is used for a refined comparison, accounting for potential temporal variations and guaranteeing the precise alignment of sequences. This comprehensive strategy ensures a balance between computational efficiency and classification accuracy, providing a robust solution for human action recognition. The steps of classification are detailed in sections \ref{sec:featurevecsec} and \ref{sec:knnclass}. 

\section{{Feature Space Design}}
\label{sec:featurevecsec}
In the realm of human pose estimation and analysis, the extraction of meaningful features from detected landmarks is of prime importance. The proposed methodology focuses on extracting embeddings from 13 key anatomical landmarks, which are represented by their 3D coordinates \((x, y, z)\).

\subsection{Landmarks Identification}

The identified landmarks in the human body are as follows: Nose, Left and Right Shoulder, Left and Right Elbow, Left and Right Wrist, Left and Right Hip, Left and Right Knee, and Left and Right Heel.

\subsection{Pose Normalization}

To ensure consistency across different poses and individuals, the detected landmarks undergo a normalization process. The normalization is performed in three stages:

\begin{enumerate}
    \item \textbf{Translation Normalization}: The landmarks are translated such that the centre of the pose (midpoint between the hips) is at the origin.
    \item \textbf{Scale Normalization}: The landmarks are scaled based on the size of the torso or the maximum distance from any landmark to the pose centre, multiplied by a given torso size multiplier.
    \item \textbf{Orientation Normalization (optional)}: The landmarks are rotated to align the vector connecting the hip centre to the shoulder centre with a predefined target direction, ensuring an upright orientation of the pose.
\end{enumerate}

\subsection{Embedding Calculation}

For the efficient extraction of features from these landmarks, an \texttt{EmbeddingCalculator} class has been utilized. 
In constructing the feature space for action classification, we prioritized features with inherent resilience to noise and occlusions, essential in 3D pose estimation. Selection focused on relative positions and orientations between joints, as these are less sensitive to occlusions and provide a stable reference in noisy data. Additionally, incorporating temporal features like joint velocity and acceleration helps smooth out noise over time. Depth information plays a crucial role in enhancing occlusion handling, allowing for a more accurate estimation of partially visible actions. This strategic selection ensures our classification remains robust across diverse and challenging conditions.

\begin{enumerate}
    \item \textbf{Single Joint Operations}: Processes features derived from individual landmarks.
    \item \textbf{Joint Pair Operations}: Processes features derived from pairs of landmarks.
    \item \textbf{Tri Joint Operations}: Processes features derived from groups of three landmarks.
\end{enumerate}
\subsubsection{Single Joint Operations}

For each landmark, the following are calculated:

\begin{itemize}
    \item \textbf{Joint Vector}: Directly takes the 3D coordinates of the landmark. Provides the spatial positioning necessary for accurate pose recognition, crucial for interpreting directional UAV commands based on limb orientation.
    \begin{equation}
        \text{Joint Vector} = \text{landmarks}[i] 
    \end{equation}
    
    where \( i \) is the index of the landmark in the predefined list.
    \item \textbf{Joint Velocity (\( v \))}: The rate of change of the joint's position with respect to time. Both velocity and acceleration are essential for distinguishing dynamic gestures from static poses, enabling the UAV to interpret the urgency or intended pace of human commands. It is calculated as: 
    \begin{equation}
        v = \frac{\text{current joint vector} - \text{previous joint vector}}{\text{current timestamp} - \text{previous timestamp}}
    \end{equation}

    \item \textbf{Joint Acceleration (\( a \))}: The rate of change of a joint's velocity with respect to time. It is computed as: 
    \begin{equation}
        a = \frac{v - \text{previous } v}{\text{current timestamp} - \text{previous timestamp}}
    \end{equation}
    \item \textbf{Joint Vector Angle}: The angle between the joint vector and each of the coordinate axes (x, y, z). Offers insights into the limb orientation, critical for understanding gesture directionality and ensuring precise UAV response to commands like vertical takeoff or horizontal movement. For a given joint vector \( \mathbf{v} \) and axis \( \mathbf{a} \), the angle \( \theta \) is computed using the dot product: 
    \begin{equation}
        \theta = \arccos \left( \frac{\mathbf{v} \cdot \mathbf{a}}{\|\mathbf{v}\| \|\mathbf{a}\|} \right)
    \end{equation}
    
    \item \textbf{Joint Angular Velocity}: The rate of change of a joint's vector angle with respect to time. Joint angular velocity and acceleration help the system gauge the smoothness or abruptness of movements, aiding in the interpretation of gesture urgency for immediate or deliberate UAV actions. 
    \item \textbf{Joint Angular Acceleration}: The rate of change of a joint's angular velocity with respect to time.
    \item \textbf{Displacement Vector}: The change in position of the joint from its previous position. Indicates the trajectory of joint movements, guiding the UAV in adjusting its flight path to align with the operator's intended direction.
\end{itemize}

\subsubsection{Joint Pair Operations}

For each pair of landmarks, the following are computed:

\begin{itemize}
    \item \textbf{Joint Pair Vector}: The difference in the 3D coordinates of the two landmarks. Provides a relational understanding of body posture by examining vectors between pairs of joints, aiding in the nuanced differentiation of gestures for accurate UAV command interpretation.
    \begin{equation}
        \text{Joint Pair Vector} = \text{landmark}[j] - \text{landmark}[k]
    \end{equation}
     where \( j \) and \( k \) are the indices of the two landmarks.
    \item \textbf{Joint Pair Velocity, Acceleration, Vector Angle, Angular Velocity, and Angular Acceleration}: These are calculated similarly to the single joint operations, but are applied to the joint pair vector.
\end{itemize}

\subsection{Tri Joint Operations}

Given the three landmarks \( A \), \( B \), and \( C \), we can define two vectors:
\begin{align}
\vec{AB} &= B - A \\
\vec{BC} &= C - B
\end{align}

\subsubsection{Tri Joint Angle}

The angle \( \theta \) between two vectors \( \vec{AB} \) and \( \vec{BC} \) is given by:
\begin{equation}
\theta = \arccos \left( \frac{\vec{AB} \cdot \vec{BC}}{\|\vec{AB}\| \|\vec{BC}\|} \right)
\end{equation}
where \( \vec{AB} \cdot \vec{BC} \) is the dot product of the two vectors.

The normal to the plane containing \( A \), \( B \), and \( C \) is given by the cross product of the vectors \( \vec{AB} \) and \( \vec{BC} \):
\begin{equation}
\vec{N} = \vec{AB} \times \vec{BC}
\end{equation}

The unit normal vector \( \hat{N} \) is then:
\begin{equation}
\hat{N} = \frac{\vec{N}}{\|\vec{N}\|}
\end{equation}

The tri joint angle \( \Theta \) (or the feature we are considering) is then a combination of \( \theta \) and \( \hat{N} \), which could be represented as:
\begin{equation}
\Theta = \hat{N} \times \theta
\end{equation}

Our selection of the feature vector was grounded in its capability to uniquely represent the anatomical structure and dynamics. It captures the geometric configuration of poses involving bends or twists, enabling complex gesture recognition for sophisticated UAV manoeuvre commands. The unit normal vector distinctively identifies the plane in which rays connecting a landmark to its neighbouring joints reside. Concurrently, the cosine of the angle effectively captures the relative positioning of these rays. By taking the product of these two entities, we obtain a singular, robust feature vector. This vector augments the feature space, bolstering our ability to discern and classify sequences with heightened precision.

\subsubsection{Tri Joint Angular Velocity}

The angular velocity \( \omega \) for the tri joint angle is the rate of change of \( \Theta \) with respect to time:
\begin{equation}
\omega = \frac{\Delta \Theta}{\Delta t}
\end{equation}

\subsubsection{Tri Joint Angular Acceleration}

The angular acceleration \( \alpha \) for the tri joint angle is the rate of change of \( \omega \) with respect to time:
\begin{equation}
\alpha = \frac{\Delta \omega}{\Delta t}
\end{equation}

\subsection{Feature Vector Extraction and Normalization}

Upon processing the normalized landmarks with the embedding calculator, we obtain the primary feature vectors, as explained in the previous sub-sections. These vectors, imbued with the dynamics of human movement, are central to our classification scheme.

To ensure a consistent representation across the dataset, we calculate certain parameters. Specifically, for each embedded sample, we ascertain its minimum (min) and maximum (max) values. These extremities are extracted from a concatenated array of embeddings, which is aggregated frame by frame from each sample.

Post parameter estimation, we normalize the feature vectors using min-max scaling. This normalization is pivotal in ensuring that no specific feature overshadows others during the classification process. By mapping the features to a range between -1 and 1, we achieve uniformity in their magnitudes while maintaining their sign. 

Integrating these features into our classification framework allows for an advanced, nuanced understanding of human motions, ensuring the UAV actions are tightly coupled with the operator's intent. This synergy between human gestures and UAV response is fundamental for applications requiring intuitive, real-time robot control.

\section{Enhanced k-Nearest Neighbor Classifier through Dimensionality Reduction}
\label{sec:knnclass}
Our innovative approach to action classification combines the strengths of dimensionality reduction via autoencoders and an augmented k-nearest Neighbor (kNN) algorithm integrated with Dynamic Time Warping (DTW). This two-step methodology is tailored for varying computational efficiency and accuracy requirements.

\subsubsection{Dimensionality Reduction with Autoencoders}

In the first pathway, we deploy a deep autoencoder for significant dimensionality reduction of the input space. The autoencoder architecture comprises a series of dense layers that form an encoding phase, transitioning from an input dimension $D_{\text{input}}$ to a reduced latent dimension $D_{\text{latent}}$, where $D_{\text{latent}} \ll D_{\text{input}}$. Formally, the encoder function $E: \mathbb{R}^{D_{\text{input}}} \rightarrow \mathbb{R}^{D_{\text{latent}}}$ compresses the data, and the decoder function $D: \mathbb{R}^{D_{\text{latent}}} \rightarrow \mathbb{R}^{D_{\text{input}}}$ aims to reconstruct the original data. The reconstruction loss is minimized, $L_{\text{reconstruction}} = \|X - D(E(X))\|_2^2$, where $X$ denotes the input data.

After dimensionality reduction, the latent representations are processed through a kNN classifier augmented with DTW as the similarity metric, enhancing accuracy and making it suitable for precision-critical scenarios.

\subsection{Dynamic Time Warping (DTW)}

Following the encoding of the candidate set \( C \), we refine our matches employing the DTW algorithm. The DTW distance between two sequences \( S \) and \( S' \) is computed as:

\begin{equation}
D_{\text{DTW}}(S, S') = \min \sum_{(i,j) \in \text{path}} d(s_i, s'_j)
\end{equation}
Here, \( d(s_i, s'_j) \) denotes the Euclidean distance between the respective feature vectors, while the "path" symbolizes the optimal alignment between the sequences.

\subsection{Classification}

Post the DTW filtering, the sequences in our final shortlist dictate the classification outcome. Given the frequency distribution of each class within the shortlist, the class exhibiting the highest prevalence is designated to the incoming sequence. Mathematically, for an incoming sequence \( S \), the assigned class \( C^* \) is:

\begin{equation}
C^* = \arg\max_{c \in \text{Classes}} \text{Frequency}(c, \text{Shortlist})
\end{equation}
The selection of \textit{k} in the kNN classifier is critical for the balance between noise sensitivity and the classifier's generalization ability. We determined the optimal \textit{k} empirically using a cross-validation approach on various dataset segments to achieve a balance that maximizes classification performance while minimizing error rates. The augmentation of kNN with Dynamic Time Warping (DTW) further enhances its sensitivity to the temporal dynamics of actions, ensuring that the dimensionality reduction does not compromise the classifier's ability to distinguish between similar movements.

\subsection{Mathematical Formulation and Empirical Evaluation}

Our methodology was empirically evaluated against the UTD-MHAD\cite{mhad} dataset to quantify the performance metrics of accuracy and computational efficiency. The autoencoder-based method focused on reconstruction loss and classification accuracy, utilizing the formula $Accuracy = \frac{TP+TN}{TP+TN+FP+FN}$, where $TP, TN, FP$, and $FN$ represent true positives, true negatives, false positives, and false negatives, respectively.

\section{UAV Control}
The UAV continuously monitors the position of the human and does so in two steps. In the first step, it tracks the operator's position on the video feed and perpetually corrects its heading, such that the human remains in the centre of its field of view. The next step, it monitors the distance of the human from the drone and tries to maintain a set distance, thereby eliminating the need for the human to stay in one place. The UAV is able to do this while simultaneously receiving commands from the human's actions and performing them.

\section{Results}

\subsection{Method Verification and Benchmarking}
\label{sec:systemverification}
To ensure the robustness and reliability of our proposed method, we initially subjected it to rigorous benchmarking. Several performance metrics were employed to verify the efficiency and accuracy of our classifier.

\noindent \textbf{Performance Metrics:}
\begin{itemize}
\item \textbf{Accuracy:} The overall rate of correctly classified human movements among all classifications.
\item \textbf{F1 Score:} A weighted average of precision and recall, providing a balance between false positives and false negatives.
\item \textbf{Confusion Matrix:} A detailed breakdown of true positives, false positives, true negatives, and false negatives.
\item \textbf{Computational Time:} The time taken for the classifier to process an input and generate an output.
\end{itemize}

The performance metrics are summarized in Table \ref{tab:performance_metrics}. These tests were run on the same Intel NUC that is present in our UAVs \cite{hert2023}, and therefore accurately reflect the ground reality. The approach was tested on the UTD-MHAD Dataset \cite{mhad}, which consists of 27 classes. We also conducted tests with a reduced number of classes to directly compare our method with those in \cite{Schneider2019}. Six classes were chosen for this comparison, with one additional class (a8) added to evaluate its effect on algorithm performance. These 7 actions were, a1: Arm swipe to the left, a6: Cross arms in the chest, a7: Basketball shoot, a8: Hand draw x, a9: Hand draw circle (clockwise), a24: Sit to stand, a27: Forward lunge.

Two versions of the proposed approach were tested on the dataset. The Heavy version utilizes the full set of encodings and is expected to be very accurate, but considerably slower. The encoded version incorporates dimensionality reduction, providing faster classification at slightly lower accuracy. This version offered a good balance between accuracy and computational time for real-time classification.
\begin{table*}[ht]
\centering
\caption{Performance metrics of the proposed method.}
\label{tab:performance_metrics}
\begin{tabular}{|c|c|c|c|c|c|c|}
\hline
Configuration        & Classes         & Test/Train cases & Accuracy (\%) & Total Time (s) & Per case Time (ms) & F1 Score \\ \hline
Heavy  & 27              & 173/688        & 98.27   & 360.44        & 2083             & 98      \\
                     & 6   & 39/153         & 98.72   & 59.03        & 1513            & 99     \\
                     & 7   & 45/178         & 95.56   & 71.71        & 1594            & 96      \\ \hline
 Encoded & 27              & 173/688        & 83.24   & 23.82         & 138             & 84      \\
                     & 6  & 39/153         & 97.44   & 1.4896          & 38              & 97      \\
                     & 7   & 45/178         & 86.67   & 1.7458          & 39              & 87      \\ \hline
                     
\end{tabular}
\end{table*}
The insights drawn from this evaluation illuminate the trade-offs between accuracy and computational demands, guiding the selection of the optimal configuration for specific application scenarios.

\textbf{Heavy Configuration: Precision at the Cost of Computational Efficiency}
The Heavy configuration (characterized by its utilization of the full set of encodings) demonstrated remarkable accuracy and F1 scores across all evaluated class groupings. It achieved a pinnacle of classification precision in the 6-class setup, with an accuracy of 98.72\% and an F1 score of 99. However, this high degree of precision comes at a significant computational cost. The total time for processing 27 classes was recorded at 360.44 seconds, with a per-case time exceeding 2000 ms. This considerable computational demand demonstrates the Heavy configuration's limited applicability in real-time or resource-constrained scenarios.

\textbf{Encoded Configuration: A Pragmatic Balance}
Emerging as the balanced contender, the Encoded configuration significantly reduces computational time without drastically compromising on accuracy. For 6 classes, it maintained an impressive accuracy of 97.44\% and an F1 score of 97, with a markedly reduced per-case time of approximately 38 ms. This configuration adeptly balances computational efficiency and precision, making it an ideal candidate for real-time applications. While notable, the dip in accuracy to 83.24\% for 27 classes still positions the Encoded configuration as a robust option, capable of handling a diverse range of movements with considerable accuracy.

\textbf{Discussion on Trade-offs and Configuration Selection}
The analysis of the three configurations highlights a fundamental trade-off between computational efficiency and accuracy. The Heavy configuration, while highly accurate, may not be feasible for real-time applications due to its significant computational demands. The Encoded configuration stands out as the optimal choice for applications requiring a harmonious balance between accuracy and computational speed, offering high performance without substantial sacrifices.
For real-time applications, the Encoded configuration's balanced performance profile makes it exceptionally suitable. It provides a viable solution that accommodates the need for quick processing times, while still maintaining a high level of accuracy. This balance is crucial for deploying efficient and responsive systems in dynamic environments where both precision and speed are essential.

\begin{table}[ht]
\centering
\scriptsize
\caption{Comparison of accuracy with state-of-the-art methods.}
\begin{tabular}{|c|c|}
\hline
\textbf{Method} & \textbf{Accuracy (\%)} \\
\hline
Classical & 60 \\
Schneider et. el. \cite{Schneider2019} & 63-76 \\
Rwigema et al. \cite{Rwigema2019} & 99.4 \\
Celebi et al. \cite{Celebi2013} & 96 \\
Proposed method (Encoded) & 86-97 \\
\hline
\end{tabular}

\end{table}

\begin{figure*}[ht]
\label{conf}
  \centering
  \begin{minipage}{\columnwidth}
    \includegraphics[width=\linewidth]{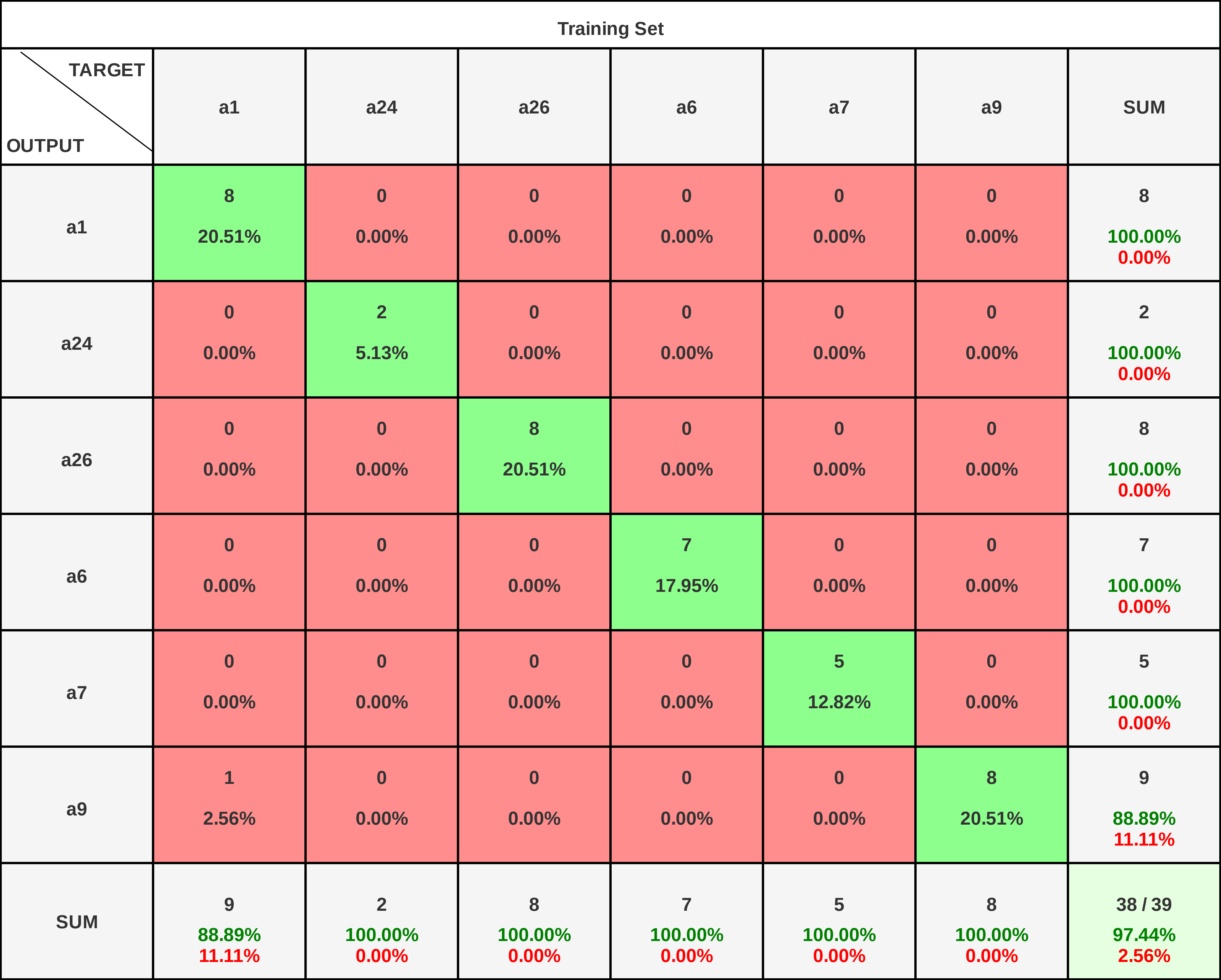}
    \caption{Confusion Matrix of Encoded Variant with 6 Gesture Classes.}
    \label{fig:conf6}
  \end{minipage}\hfill
  \begin{minipage}{\columnwidth}
    \includegraphics[width=\linewidth]{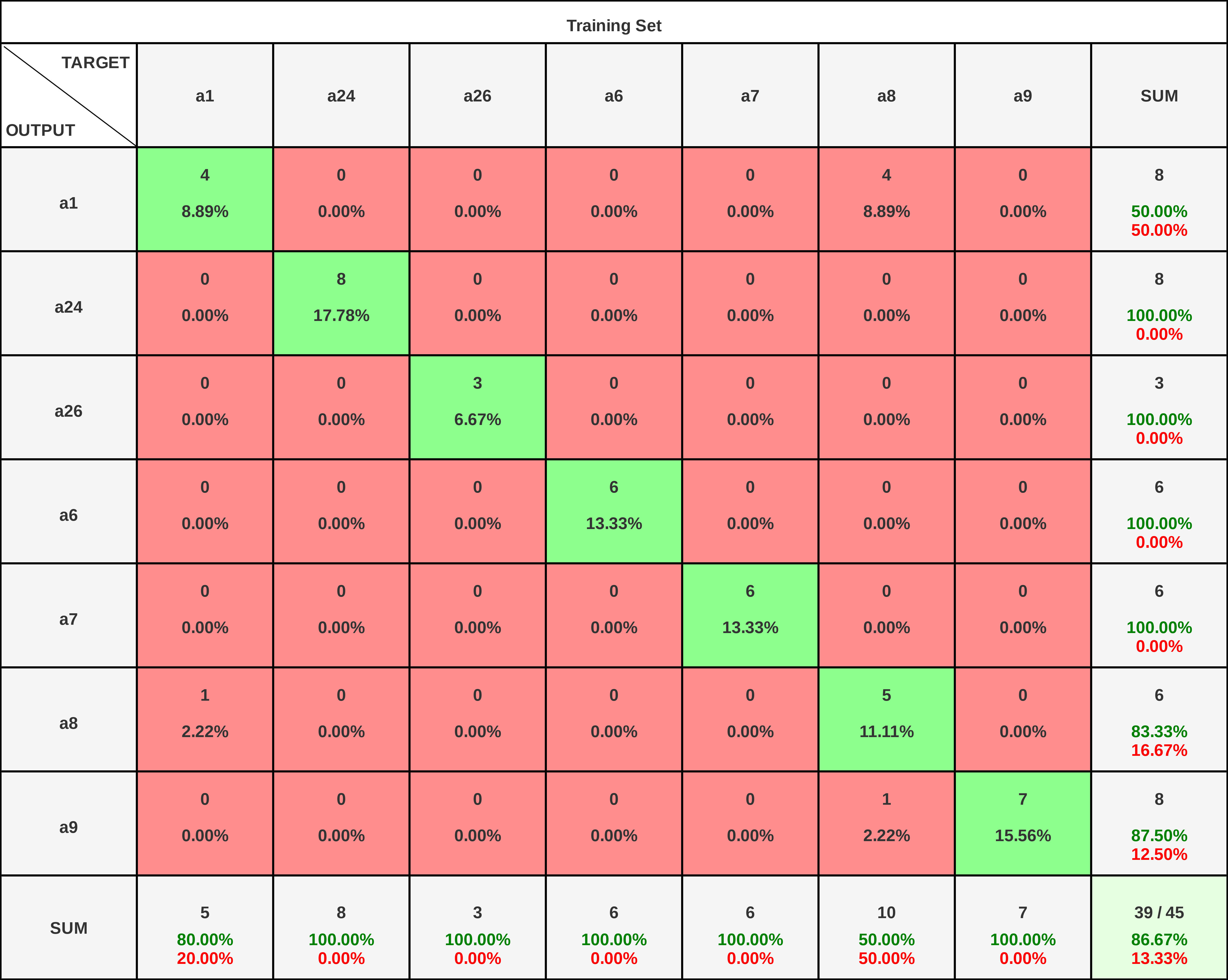}
    \caption{Confusion Matrix of Encoded Variant with 7 Gesture Classes.}
    \label{fig:conf7}
  \end{minipage}
\end{figure*}

Additionally, the confusion matrices of the Encoded variant of the approach for 6 and 7 gestures respectively are displayed in Fig \ref{fig:conf6} and Fig \ref{fig:conf7}. The model exhibits a high degree of accuracy for gestures, such as a24\_sit\_to\_stand, a26\_lunge, and a7\_basketball\_shoot, which are likely to have distinct starting and ending poses or unique motion patterns that are easily distinguishable.  Confusions are primarily seen with a1\_swipe\_left which is sometimes mistaken for a8\_draw\_X and vice versa, suggesting that the horizontal component of the swipe is similar to part of the "draw X" motion. Similarly, misclassifications between a1\_swipe\_left and a9\_draw\_circle\_cw imply that certain segments of the swipe and circular gestures may be indistinguishable to the model. The model’s difficulty in differentiating between a1\_swipe\_left and gestures involving complex hand trajectories (a8\_draw\_X, a9\_draw\_circle\_cw) indicates a potential area for improvement. Refinement of the feature set and the inclusion of more granular temporal data could enhance the model's ability to discern between these gestures with overlapping features. 

\subsection{Comparative Analysis with State-of-the-Art Methods}
In this section, we juxtapose the performance of our proposed Encoded configuration against various state-of-the-art methods, as summarized in the accompanying table. The comparative analysis is crucial to positioning our work within the broader landscape of action classification methodologies, emphasizing its competitive advantages and identifying areas for further refinement.

\textbf{Overview of Comparative Performance}
\begin{enumerate}
    \item Classical Methods: These approaches, typically involving hand-engineered features and classical machine learning algorithms, show a base accuracy of 60\%. Our method significantly surpasses this benchmark, demonstrating the efficacy of modern, data-driven approaches in handling complex classification tasks.
    \item Schneider et al.: With accuracies ranging between 63\% to 76\%, the work by Schneider et al. closely aligns with the initial performance metrics our study aimed to exceed. By achieving accuracy between 86\% to 97\% in the Encoded configuration, our method not only surpasses Schneider et al.'s performance, but also showcases the potential of embedding calculators and dimensionality reduction techniques in enhancing classification accuracy.
    \item Rwigema et al.: Although Rwigema et al.'s method achieves an impressive accuracy of 99.4\%, it is noted for its unsuitability for real-time applications due to substantial computational requirements. This highlights a critical aspect of our research focus—balancing high accuracy with computational efficiency to enable real-time classification.
    \item Celebi et al.: The method by Celebi et al. presents a high accuracy of 96\%, situating it as a leading approach within the field. Our proposed method's performance falls within this high-accuracy bracket while emphasizing real-time applicability.

\end{enumerate}

\subsection{Real-World Deployment}
\label{sec:realworld}
Upon verification, we proceeded to deploy our system on our UAV platform\cite{hert2022}\cite{hert2023} and our UAV control system\cite{Baca2021}, to assess its real-world applicability. The primary focus of this phase was to determine how well our classifier could translate laboratory results into practical, actionable commands in an outdoor environment.

\begin{figure}[!t]
\vspace{0.3cm}
\centerline{\includegraphics[width = 0.45\textwidth]{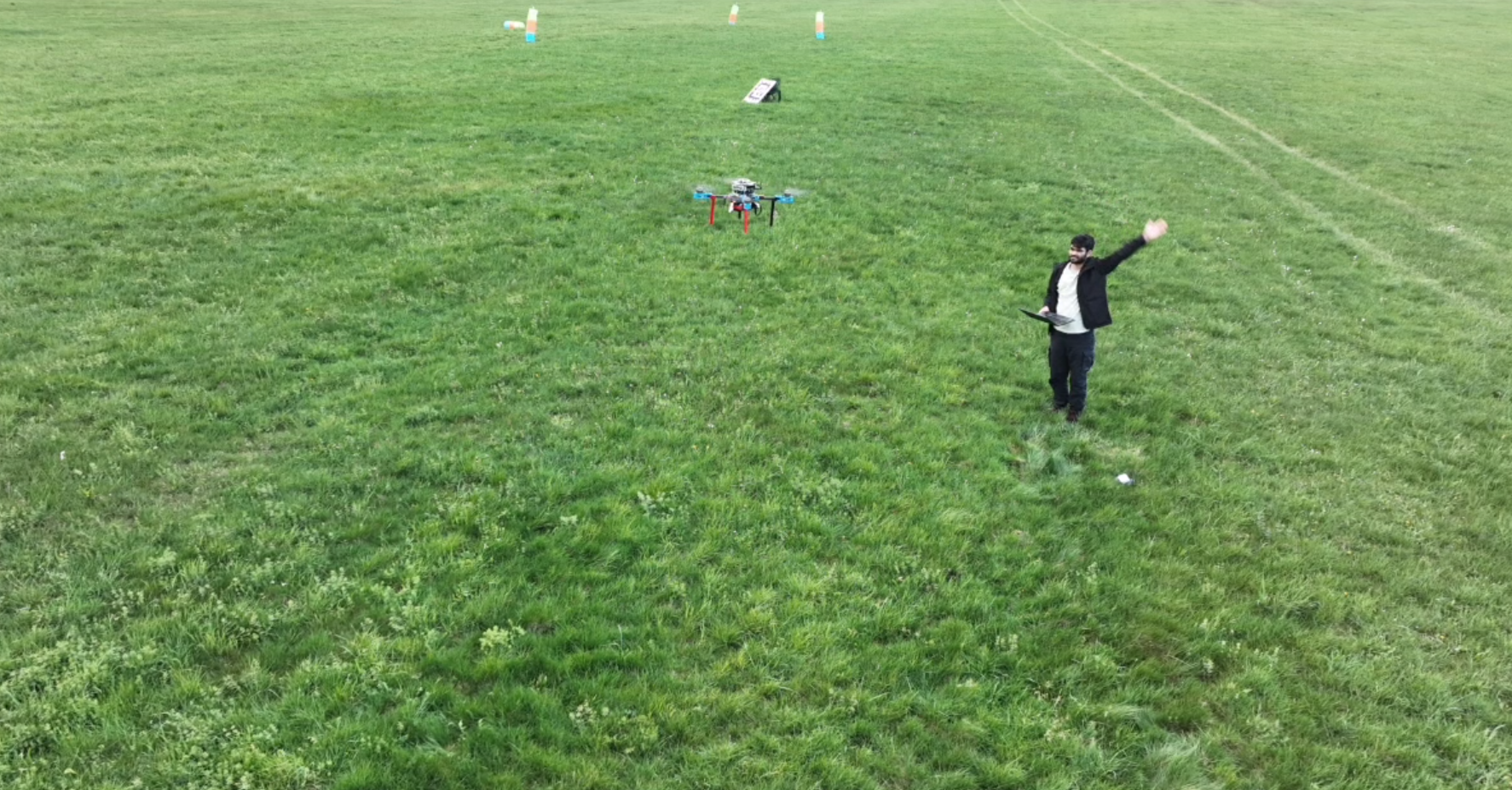}}
\caption{A UAV being controlled by a human operator in an open field.}
\label{fig:result}
\end{figure}

During the trials, the UAV was subjected to a series of predefined human actions. The 6 gestures that were used for lab validation (a1, a6, a7, a9, a24, a26), were performed for testing, with the human standing between 4-8 meters away from the UAV. The UAV correctly recognized and responded to 19 out of 20 actions, yielding a real-world accuracy rate of 95\%. Notably, the proposed approach adeptly handled dynamic environmental factors, such as changing light conditions and background noise due to the presence of clouds, occluding sunlight sporadically throughout the experiment, showcasing its adaptability and robustness. Additionally, we encountered no false positives, which is crucial as the performance of unintended actions is undesirable. The one action that was not correctly identified was classified as a null action, leading to no command being sent to the drone. This is the intended behaviour that we want our approach to adopt. Missing an action is preferable to misidentifying an action and behaving erratically. Fig \ref{fig:result} shows a UAV being controlled by a human operator using gestures. 

\section{Conclusion}
In conclusion, the feature space design offers a comprehensive approach to extracting rich embeddings from human pose landmarks. These embeddings, grounded in both anatomical significance and mathematical rigour, are poised to enhance the capabilities of pose-based analysis systems. The comparative analysis underscores the Encoded configuration as the preferred choice for a wide range of applications, particularly those necessitating real-time processing. It embodies a practical compromise, delivering high accuracy and F1 scores with considerably lower computational times compared to the Heavy configuration. The comparative analysis also elucidates the positioning of our proposed method within the action classification domain. By offering a substantial improvement over classical methods and some contemporary approaches, as well as by providing a viable alternative to high-accuracy, computationally intensive methods, our work carves out a niche in the pursuit of real-time, efficient, and accurate action classification. It underscores the importance of methodological advancements that do not sacrifice practical applicability for theoretical precision, thereby aligning with the evolving needs of real-world applications. The proposed method performed exceptionally when deployed on a real UAV, proving its capability in real-world applications.



\bibliographystyle{apalike}
{\small
\bibliography{root}}

\end{document}